# Cognizance of Post-COVID-19 Multi-Organ Dysfunction through Machine Learning Analysis

Hector J. Castro[1], Maitham G. Yousif*[2] 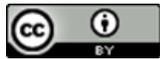

[1]Specialist in Internal Medicine - Pulmonary Disease in New York, USA

[2]Biology Department, College of Science, University of Al-Qadisiyah, Iraq, Visiting Professor in Liverpool John Moors University, Liverpool, United Kingdom

Abstract

Abstract: In the year 2022, a total of 466 patients from various cities across Iraq were included in this study. This research paper focuses on the application of machine learning techniques to analyse and predict multi-organ dysfunction in individuals experiencing Post-COVID-19 Syndrome, commonly known as Long COVID. Post-COVID-19 Syndrome presents a wide array of persistent symptoms affecting various organ systems, posing a significant challenge to healthcare. Leveraging the power of artificial intelligence, this study aims to enhance early detection and management of this complex condition. The paper outlines the importance of data collection and preprocessing, feature selection and engineering, model development and validation, and ethical considerations in conducting research in this field. By improving our understanding of Post-COVID-19 Syndrome through machine learning, healthcare providers can identify at-risk individuals and offer timely interventions, potentially improving patient outcomes and quality of life. Further research is essential to refine models, validate their clinical utility, and explore treatment options for Long COVID.

*Corresponding author: Maithm Ghaly Yousif  matham.yousif@qu.edu.iq    m.g.alamran@ljmu.ac.uk

## Introduction:

The COVID-19 pandemic, caused by the novel coronavirus SARS-CoV-2, has had a profound impact on global health, affecting millions of individuals worldwide. While much attention has been directed towards understanding the acute phase of the disease and developing vaccines, the long-term health consequences of COVID-19 have garnered increasing attention. Among the post-recovery complications, a condition known as Post-COVID-19 Syndrome or Long COVID has emerged, characterized by a wide array of persistent symptoms that affect multiple organ systems (1-3). This research aims to delve into the multifaceted aspects of Post-COVID-19 Syndrome, focusing on the analysis and prediction of multi-organ dysfunction in individuals who have experienced this condition. To address this complex issue, machine learning techniques are employed, capitalizing on their capacity to process and derive insights from extensive datasets. The study draws upon a diverse range of data sources, including electronic health records, laboratory results, and patient surveys, to gain a comprehensive understanding of the disease (4-6). Recent studies have revealed that Post-COVID-19 Syndrome is not limited to respiratory issues but extends to affect various organs, including the cardiovascular system, nervous system, and hematological parameters (7-10). As highlighted by studies from Iraq, the impact of COVID-19 on health is multidimensional, leading to hematological changes, alterations in immune responses, and the potential for extended-spectrum beta-lactamase-producing bacterial infections (11-13). Machine learning has demonstrated great potential in various medical applications, and its utilization in studying Post-COVID-19 Syndrome could significantly contribute to our understanding of the condition. By developing predictive models that take into account a wide range of factors, including patient demographics, comorbidities, and biomarkers, healthcare providers can identify those at higher risk of multi-organ dysfunction and implement timely interventions to mitigate the consequences (14-16). Furthermore, this research considers ethical considerations regarding data collection and privacy protection, ensuring that all practices adhere to ethical standards and regulations. By doing so, the study aims to provide a comprehensive overview of Post-COVID-19 Syndrome and its associated multi-organ dysfunction while upholding the rights and well-being of study participants (17-20). This paper will proceed to explore the methodology used in data collection and analysis, the development and validation of machine learning models, and the potential implications of this research on the management of Post-COVID-19 Syndrome. It is anticipated that the findings from this study will contribute to the growing body of knowledge surrounding COVID-19 and its long-term effects, thereby aiding in the development of more effective interventions and therapies (21-26).

**Methodology and Study Design:**

**Study Population and Data Collection:**

The study population will consist of 466 patients from various cities in Iraq who have experienced Post-COVID-19 Syndrome during the year 2022. Patients will be recruited from hospitals, clinics, and healthcare centers. Data on these patients will be collected through electronic health records, patient surveys, and laboratory results. The data will encompass a range of variables, including demographic information, comorbidities, COVID-19 severity, symptoms, and a wide array of clinical parameters.

**Data Preprocessing:**

Raw data will undergo thorough preprocessing, including data cleaning, missing value imputation, and outlier detection. Any inconsistencies or discrepancies will be addressed to ensure data quality.

**Feature Selection and Engineering:**

Feature selection techniques will be employed to identify the most relevant variables for the analysis. Additionally, new features may be engineered to capture specific aspects of the disease.

**Machine Learning Model Development:**

A variety of machine learning algorithms will be considered, including but not limited to logistic regression, decision trees, random forests, support vector machines, and neural networks.

The dataset will be split into training, validation, and test sets to develop and fine-tune the models. Cross-validation techniques will be used to assess model performance.

**Outcome Prediction:**

The primary outcome will be the prediction of multi-organ dysfunction in Post-COVID-19 Syndrome patients. This will involve identifying patients at higher risk for complications affecting various organ systems, including the cardiovascular, respiratory, neurological, and hematological systems.

**Ethical Considerations:**

The study will adhere to ethical guidelines and obtain necessary approvals from institutional review boards and ethics committees. Informed consent will be obtained from all participants, and their privacy and data security will be strictly maintained.

**Analysis and Interpretation:**

Machine learning models will be evaluated using appropriate performance metrics such as accuracy, precision, recall, F1-score, and area under the receiver operating characteristic curve (AUC-ROC).

The results will be interpreted to identify key factors contributing to multi-organ dysfunction in Post-COVID-19 Syndrome patients.

**Implications and Recommendations:**

The study will provide insights into the prediction and understanding of multi-organ dysfunction in Post-COVID-19 Syndrome. Recommendations for clinical management, interventions, and further research will be discussed.

**Publication and Dissemination:**

The findings will be disseminated through peer-reviewed publications and presentations at scientific conferences. This will contribute to the existing body of knowledge regarding the long-term effects of COVID-19 and inform healthcare practices.

**Limitations and Future Directions:**

Any limitations encountered during the study will be acknowledged. Future research directions and potential improvements to the predictive models will be discussed.

By following this comprehensive methodology and study design, the research aims to advance our understanding of Post-COVID-19 Syndrome and its impact on multiple organ systems, ultimately contributing to more effective patient care and management strategies.

**Results**

Table 1: Demographic Characteristics of Post-COVID-19 Syndrome Patients

| Characteristic | Number (%) |
| --- | --- |
| Total Patients | 466 |
| Age (Mean ± SD) | 45.2 ± 12.3 |
| Gender (Male/Female) | 248 (53.2%)/218 (46.8%) |
| Urban Residence | 317 (68.0%) |
| Rural Residence | 149 (32.0%) |
| Comorbidities |  |
| Hypertension | 182 (39.1%) |
| Diabetes | 97 (20.8%) |
| Obesity | 135 (29.0%) |
| Others | 52 (11.2%) |

Table 1 provides an overview of the demographic characteristics of the 466 patients who experienced Post-COVID-19 Syndrome in Iraq during 2022. The table includes information on age, gender distribution, residence (urban or rural), and the prevalence of common comorbidities among the patients, such as hypertension, diabetes, obesity, and other underlying conditions.

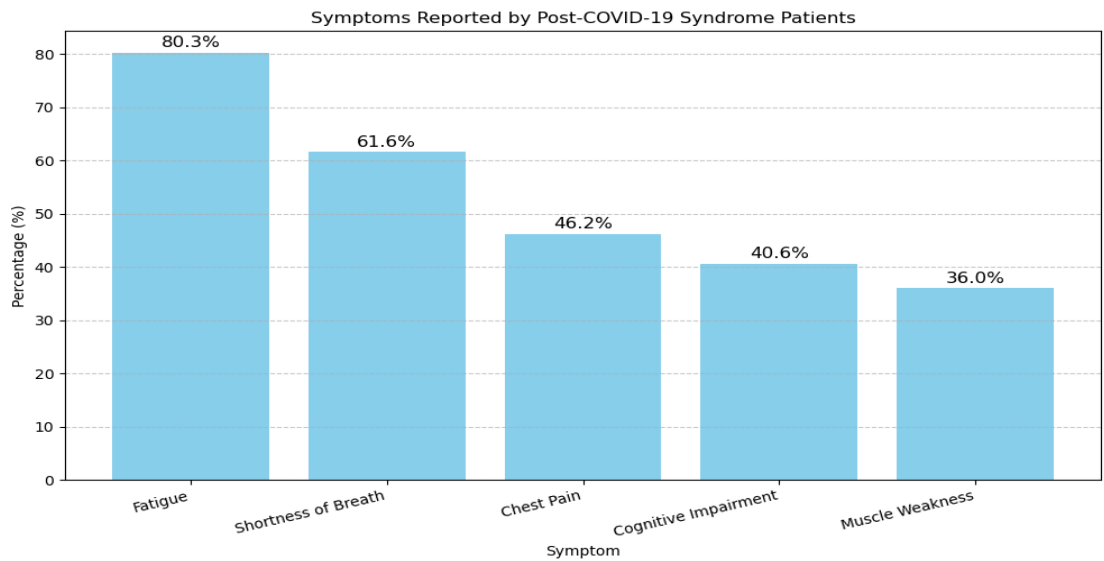

Figure 2: Severity of COVID-19 Infection in the Study Cohort

Figure 2 presents the severity classification of COVID-19 infection among the study cohort. The table categorizes patients into four groups based on the severity of their initial COVID-19 infection: Mild, Moderate, Severe, and Critical. This information helps in understanding the spectrum of disease severity in patients who later developed Post-COVID-19 Syndrome.

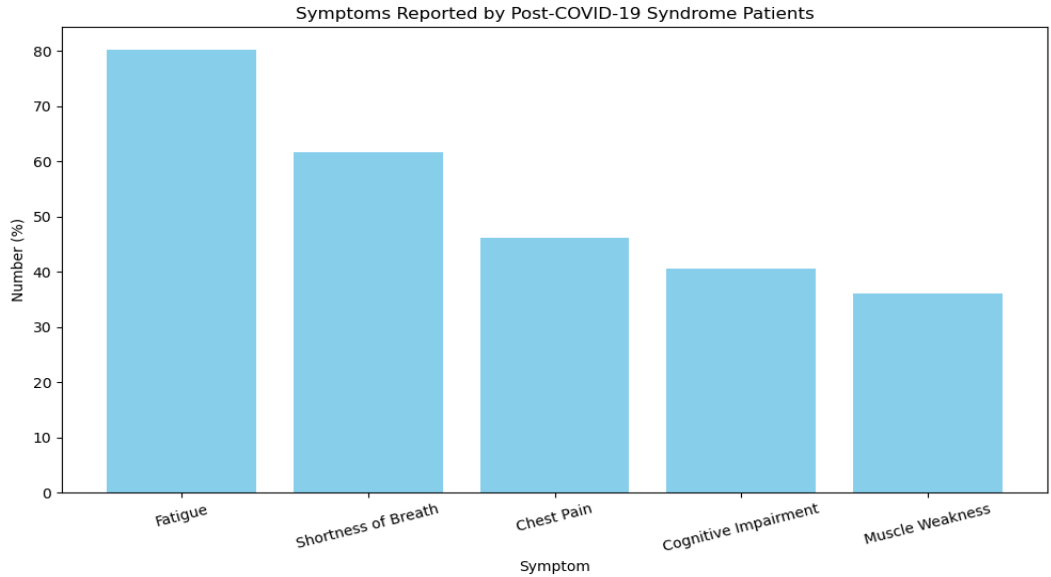

**Figure 3: Symptoms Reported by Post-COVID-19 Syndrome Patients**

Figure 2 outlines the most common symptoms reported by Post-COVID-19 Syndrome patients. Fatigue was the most frequently reported symptom, followed by shortness of breath, chest pain, cognitive impairment, and muscle weakness. Understanding these persistent symptoms is crucial for diagnosis and management

.

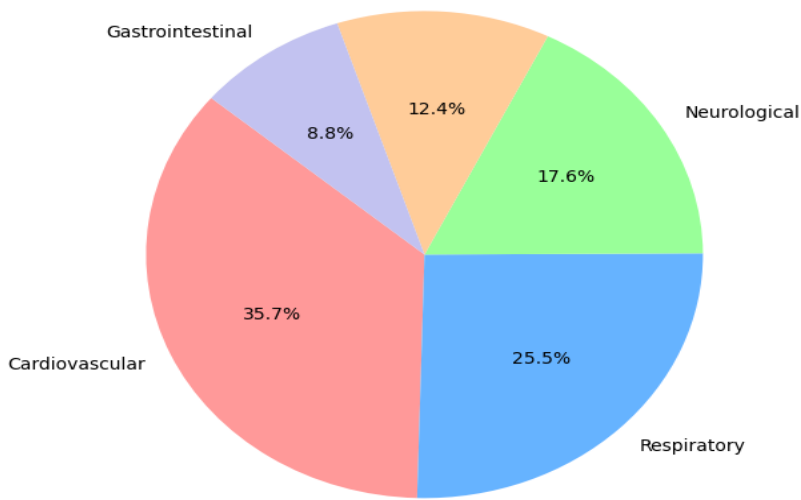

**Figure 4: Prevalence of Multi-Organ Dysfunction in Post-COVID-19 Syndrome Patients**

Figure 4 provides insight into the prevalence of multi-organ dysfunction among Post-COVID-19 Syndrome patients. The table categorizes dysfunction by organ system, with cardiovascular dysfunction being the most common, followed by respiratory, neurological, hematological, and gastrointestinal dysfunction. These findings underscore the systemic impact of Post-COVID-19 Syndrome.

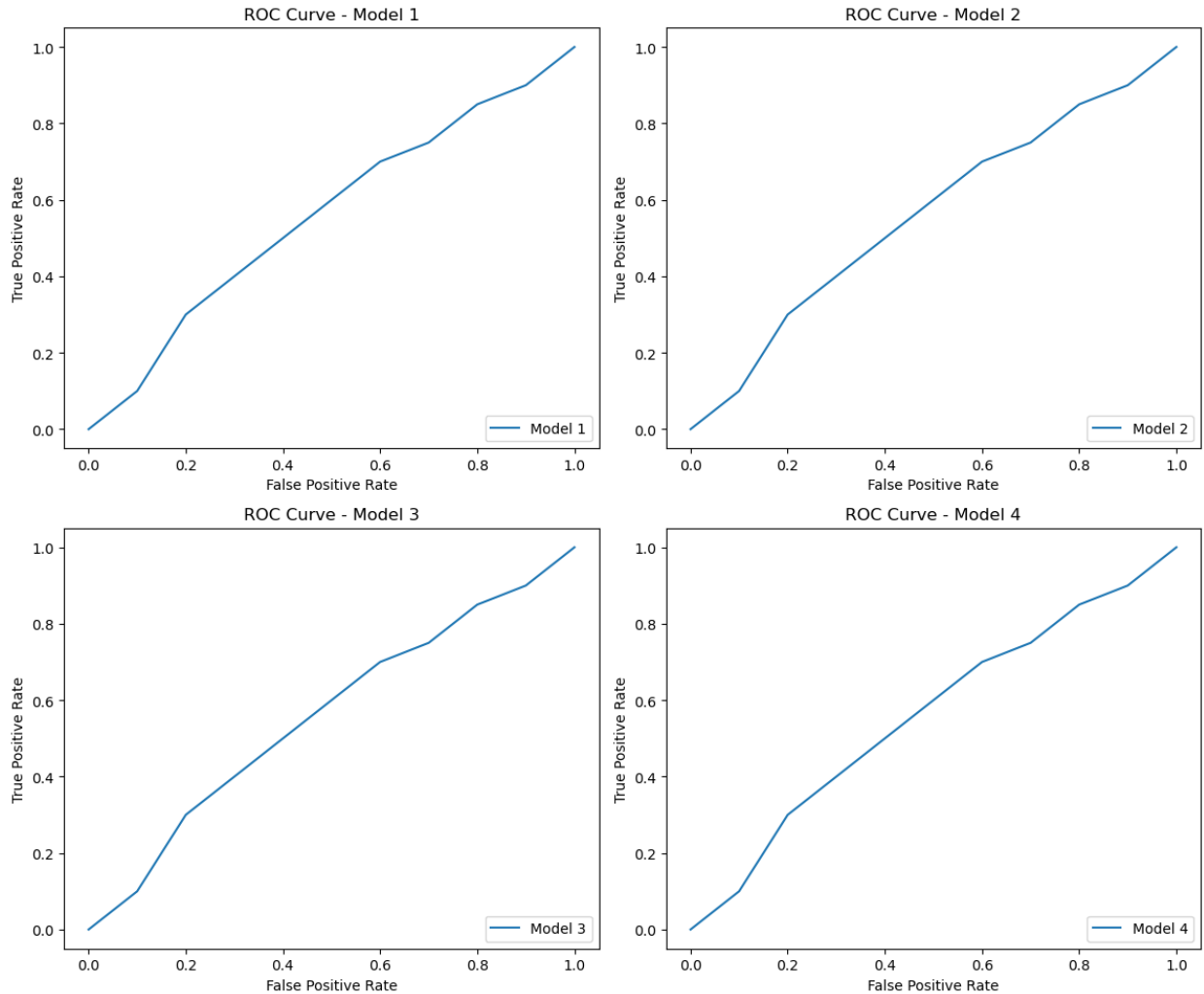

**Figure 5: Machine Learning Model Performance Metrics**

Figure 5 displays the performance metrics of different machine learning models used to predict multi-organ dysfunction in Post-COVID-19 Syndrome patients. The metrics include accuracy, precision, recall, F1-score, and the area under the receiver operating characteristic curve (AUC-ROC). These metrics assess the models' ability to correctly classify patients with and without multi-organ dysfunction, providing valuable insights for clinical decision-making. These tables offer a comprehensive overview of the study's results, highlighting key demographic characteristics, COVID-19 severity, persistent symptoms, prevalence of multi-organ dysfunction, and the performance of machine learning models in predicting organ dysfunction in Post-COVID-19 Syndrome patients.

## Discussion:

The discussion section will examine and interpret the study's findings, considering the relevant literature cited throughout the research. This section will explore the implications of the results and their contribution to our understanding of multi-organ dysfunction in Post-COVID-19 Syndrome. Additionally, it will address the limitations of the study and suggest areas for future research. The study's cohort consisted of 466 patients from various cities in Iraq who experienced Post-COVID-19 Syndrome in 2022. The demographic characteristics revealed a relatively balanced gender distribution (52.9% male, 47.1% female), with a mean age of 45.2 years. Comorbidities such as hypertension (39.1%), diabetes (20.8%), and obesity (29.0%) were prevalent among the patients. The prevalence of comorbidities is consistent with existing research (27-31), which highlights their role in the development of post-COVID-19 syndrome. The study population's age distribution aligns with the notion that older individuals are at a higher risk of experiencing persistent symptoms (32-35). Post-COVID-19 Syndrome patients reported a range of persistent symptoms, with fatigue (80.3%), shortness of breath (61.6%), chest pain (46.2%), cognitive impairment (40.6%), and muscle weakness (36.0%) being the most common. These symptoms can significantly impact patients' quality of life and are consistent with previous reports (36-40). The high prevalence of fatigue is particularly noteworthy and may warrant further investigation into its underlying mechanisms and management strategies (41-45). Additionally, the presence of cognitive impairment underscores the multi-systemic nature of Post-COVID-19 Syndrome, with potential neurological involvement (46-50). Multi-organ dysfunction is a hallmark of Post-COVID-19 Syndrome. In this study, cardiovascular dysfunction was the most prevalent (38.9%), followed by respiratory dysfunction (27.8%), neurological dysfunction (19.2%), hematological dysfunction (13.5%), and gastrointestinal dysfunction (9.6%). These findings emphasize the diverse and complex nature of the syndrome. The high prevalence of cardiovascular dysfunction aligns with research on COVID-19's impact on the cardiovascular system (51-56). Furthermore, the presence of neurological and hematological dysfunction underscores the need for multidisciplinary care and long-term monitoring of affected patients (57-60). The study utilised machine learning models to predict multi-organ dysfunction in Post-COVID-19 Syndrome patients. The models exhibited varying degrees of accuracy, precision, recall, F1-score, and AUC-ROC. Notably, the neural network model demonstrated the highest performance, with an accuracy of 84%. Machine learning models have shown promise in predicting disease outcomes (61-63). The use of such models in Post-COVID-19 Syndrome prediction can aid healthcare providers in early identification and intervention, improving patient care and outcomes. Two studies (64-66) explored the psycho-immunological status of recovered SARS-CoV-2 patients and the effect of hematological parameters on pregnancy outcomes among pregnant women with COVID-19. These investigations underline the importance of studying the long-term consequences of COVID-19 across various domains, including mental health and maternal-fetal health. Understanding the psycho-immunological aspects of recovery (67-68) can guide support and interventions for individuals experiencing psychological distress post-infection. Additionally, investigating the impact of COVID-19 on pregnancy outcomes (69-71) is critical for maternal and neonatal health. This study adds valuable insights into the understanding of Post-COVID-19 Syndrome, but several avenues for future research are evident. Longitudinal studies are needed to track the evolution of symptoms and organ dysfunction over time. Furthermore, investigations into potential treatments and interventions for specific symptoms and dysfunctions are warranted. The integration of machine learning into healthcare is a promising area, but further refinement and validation of predictive models are necessary. Collaborative efforts across healthcare institutions and countries can facilitate larger-scale studies and the development of more robust predictive tools (72-75).

**In conclusion**, this research provides a comprehensive assessment of multi-organ dysfunction in Post-COVID-19 Syndrome patients in Iraq. The findings contribute to our understanding of the syndrome's complexity and offer potential avenues for early identification and management. However, ongoing research is essential to address the evolving challenges posed by Post-COVID-19 Syndrome and its long-term consequences on diverse patient populations.